\documentclass[letterpaper, 10 pt, conference]{ieeeconf}  

\IEEEoverridecommandlockouts                              

\usepackage{graphics} 
\usepackage{subfig}
\usepackage{epsfig} 
\usepackage{mathptmx} 
\usepackage{times} 
\usepackage{amsmath} 
\usepackage{amssymb}  
\usepackage{color}
\usepackage{soul}
\usepackage[normalem]{ulem}
\usepackage{lipsum}
\usepackage[disable]{todonotes}

\newcommand{\comment}[1] {} 
\usepackage{amsmath} 
\usepackage{adjustbox}
\usepackage{multirow}
\usepackage{comment}
\usepackage[colorlinks,linkcolor=black]{hyperref}

\usepackage{algorithm}
\usepackage[noend]{algpseudocode}

\usepackage{array}

\usepackage{parselines}

\title{\LARGE \bf Hierarchical Reinforcement Learning Method for Autonomous Vehicle Behavior Planning$^*$
}

\author{Zhiqian Qiao$^{1}$, Zachariah Tyree$^{2}$, Priyantha Mudalige$^{2}$, Jeff Schneider$^{3}$ and John M. Dolan$^{3}$ 
\thanks{*This work is supported by General Motors}
\thanks{$^{1}$Zhiqian Qiao is Ph.D. student of Electrical and Computer Engineering, Carnegie Mellon University, 5000 Forbes Ave, Pittsburgh, USA
        {\tt\small zhiqianq@andrew.cmu.edu}}%
\thanks{$^{2}$ Research \& Development, General Motors}
\thanks{$^{3}$ Faculties of The Robotics Institute, Carnegie Mellon University}
}

\begin{document}

\maketitle
\thispagestyle{empty}
\pagestyle{empty}

\begin{abstract}

   In this work, we propose a hierarchical reinforcement learning (HRL) structure which is capable of performing autonomous vehicle planning tasks in simulated environments with multiple sub-goals. In this hierarchical structure, the network is capable of 1) learning one task with multiple sub-goals simultaneously; 2) extracting attentions of states according to changing sub-goals during the learning process; 3) reusing the well-trained network of sub-goals for other similar tasks with the same sub-goals. The states are defined as processed observations which are transmitted from the perception system of the autonomous vehicle. A hybrid reward mechanism is designed for different hierarchical layers in the proposed HRL structure. Compared to traditional RL methods, our algorithm is more sample-efficient since its modular design allows reusing the policies of sub-goals across similar tasks. The results show that the proposed method converges to an optimal policy faster than traditional RL methods.

\end{abstract}

\section{INTRODUCTION}

In a traditional autonomous vehicle (AV) system, after receiving the processed observations coming from the perception system, the ego vehicle performs behavior planning to deal with different scenarios or environments. At the behavior planning level, algorithms generate high-level decisions such as \textit{Go}, \textit{Stop}, \textit{Follow front vehicle}, etc. After that, a lower-level trajectory planning system maps those high-level decisions to trajectories according to map and dynamic object information. Then a lower-level controller outputs the detailed pedal or brake inputs to allow the vehicle to follow these trajectories.

At first glance, among algorithms generating behavior decisions, rule-based algorithms \cite{carfollow}\cite{ttc} appear to describe human-like decision processes well. However, estimating  other vehicles' behaviors accurately and adjusting the corresponding decisions to account for changes in the environment is difficult if the decisions of the ego car are completely hand-engineered. This is because the environment can vary across many different dimensions, all relevant to the task of driving, and the number of rules necessary for planning in this nuanced setting can be untenable.

An alternative method is reinforcement learning \cite{dqn}\cite{ddpg}\cite{go}. In recent works, RL has been used to solve some particular problems by designing states, actions and reward functions in a simulated environment. For example, the related applications within the autonomous vehicle domain include learning an output controller for lane-following, merging into a roundabout, traversing an intersection and lane changing. However, low stability and large computational requirements make RL difficult to use widely for more general tasks with multiple sub-goals. Obviously, applying RL to learn the behavior planning system from scratch not only increases the difficulties of adding or deleting sub-functions within the existing behavior planning system, but also makes it harder to debug problems. A hierarchical structure which is structurally similar to the heuristic-based algorithms is more feasible and can save computation time by learning different functions or tasks separately.

\begin{figure}[!t]
  \centering
  \includegraphics[width=\columnwidth]{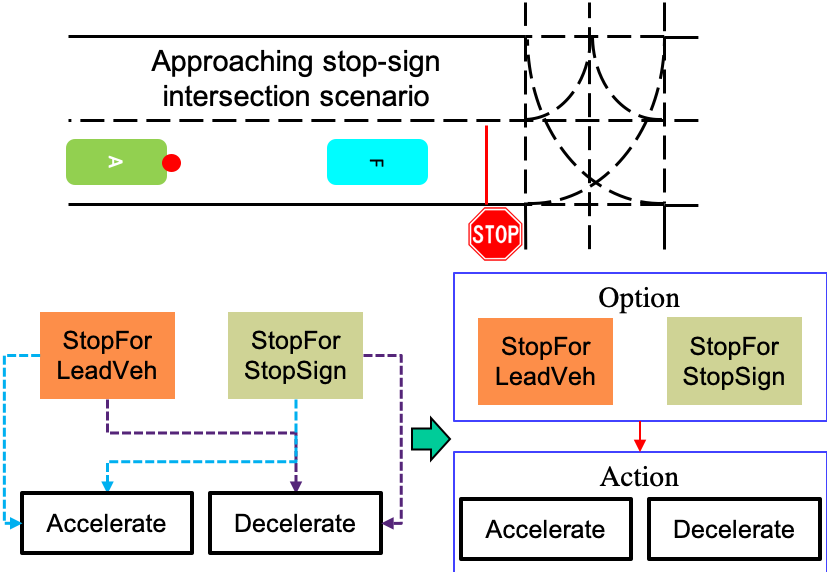}
  \caption{Heuristic-based structure vs. HRL-based structure}
  \label{fig_rule}
\end{figure}

Reinforcement learning (RL) has proven the capability of solving for the optimal policy, which can map various observations to corresponding actions in complicated scenarios. In traditional RL approaches it is often necessary to train a unique policy for each task the agent may be faced with. In order to solve a new task the entire policy must be relearned regardless of how similar the two tasks may be. Our goal in this work is to construct a single planning algorithm based on hierarchical deep reinforcement learning (HRL) which can accomplish behavior planning in an environment where the agent must pursue multiple sub-goals and to do so in a way in which any sub-goal policies can be reused for subsequent tasks in a modular fashion (see Figure \ref{fig_rule}). The main contributions of the work are:

\begin{itemize}
    \item A state attention model-based HRL structure.
    \item A hybrid reward function mechanism which can efficiently evaluate the performance among actions of different hierarchical levels.
    \item A hierarchical prioritized experience replay designed for HRL.
\end{itemize}

\section{Related Work}

This section introduces previous work related to this paper, which can be categorized as follows: 1) papers that address reinforcement learning (RL) and hierarchical reinforcement learning algorithms; 2) papers that propose self-driving behavior planning algorithms.

\subsection{Reinforcement Learning}
Based on the context of reinforcement learning, algorithms with extended functions based on RL and HRL have been proposed. \cite{hdrl} proposed the idea of a meta controller, which is used to define a policy governing when the lower-level action policy is initialized and terminated. \cite{maxq} introduced the concept of hierarchical Q learning called MAXQ, which proved the convergence of MAXQ mathematically and could be computed faster than the original Q learning experimentally. \cite{hrl} proposed an improved MAXQ method by combining the R-MAX \cite{rmax} algorithm with MAXQ. It has both the efficient model-based exploration of R-MAX and the opportunities for abstraction provided by the MAXQ framework. \cite{parl} used the idea of the hierarchical model and transferred it into parameterized action representations. They use a DRL algorithm to train high-level parameterized actions and low-level actions together in order to get more stable results than by getting the continuous actions directly.

\subsection{Behavior Planning of Autonomous Vehicles}

Previous work applied heuristic-based and learning-based algorithms to the behavior planning of autonomous vehicles based on different scenarios. For example, \cite{slot} proposed a slot-based approach to check if a situation is safe to merge into lanes or across an intersection with moving traffic. This method is based on information on slots available for merging behavior, which may include the size of the slot in the target lane, and the distance between the ego-vehicle and front vehicle. Time-to-collision (TTC) \cite{ttc} is a heuristic-based algorithm which has normally been applied in intersection scenarios as a baseline algorithm. Fuzzy logic is also a very popular heuristic-based approach to model the decision making and behavior planning for autonomous vehicles. In contrast to the vanilla heuristic-based algorithm, fuzzy logic allows adding the uncertainty of the results into the decision process. \cite{fuzzy} used a fuzzy logic method to control the traffic flow in urban intersection scenarios, where the vehicles have access to the environment information via the vehicle to vehicle (V2V) system. However, the V2V system has only been applied to a small number of public roads and few vehicle manufacturers have added V2V function into their vehicles. In \cite{fuzzyroundabouts}, the researchers developed a fuzzy logic method for the application of steering control in roundabout scenarios. 

The heuristic-based algorithms need much work from human beings to design various rules in order to deal with different scenarios in urban environments. As a result, learning-based algorithms, especially  reinforcement learning, has been applied to transfer multiple rules into a mapping function or one neural network. \cite{pomdpdecision} formulated the decision-making problem for autonomous vehicles under uncertain environments as a POMDP and trained out a Bayesian Network representation to deal with a T-shape intersection merging problem. \cite{dorsaplanning} modeled the interaction between autonomous vehicles and human drivers by the method of  Inverse Reinforcement Learning (IRL) \cite{irl} in a simulated environment. The work simulated autonomous vehicles to motivate human drivers' reactions and acquired reward functions in order to plan better decisions while controlling autonomous vehicles. \cite{transferdqn} dealt with the traversing problem via Deep Q-Networks combined with a long-term memory component. They trained a state-action function Q to allow an autonomous vehicle to traverse intersections with moving traffic. \cite{navigateintersection} used Deep Recurrent Q-network (DRQN) with states from a bird's-eye view of the intersection to learn a policy for traversing the intersection. \cite{occludednavigating} proposed an efficient strategy to navigate through intersections with occlusion by using the DRL method. Their results showed better performance compared to some heuristic methods. 

In our work, the main idea is to combine the heuristic-based decision-making structured with the HRL-based approaches in order to integrate the advantages coming from both methods. We built the HRL-structure according to the heuristic method (see Figure \ref{fig_rule}) so that the system is easier for validating different functions in the system instead of a whole neural-network black-box. 

\section{Preliminaries}

In this section, the preliminary background of the problem is described. The fundamental algorithms including Deep Q-Learing \cite{dqn}, Double Deep Q-Learning \cite{doubledqn} and Hierarchical Deep Reinforcement Learning \cite{hdrl} (HRL) are introduced in this part.

\subsubsection{Deep Q-learning and Double Deep Q-learning}

Since proposed, Deep Q-Networks and Double Deep Q-Networks have been widely applied in reinforcement learning problems. In Q-learning, an action-value function $Q_{\pi}(s, a )$ is learned to get the optimal policy $\pi$ which can maximize the action-value function $Q^*(s, a)$. Hence, a parameterized action-value function $Q(s, a | \mathbf{\theta})$ is used with a discount factor $\gamma$, as in Equation \ref{equ_q}.
\begin{equation}
    \label{equ_q}
    \begin{split}
        Q^*(s, a) & = \max_{\mathbf{\theta}} Q(s, a | \mathbf{\theta})\\
        & = r + \gamma \max_{\mathbf{\theta}} Q(s', a' | \theta)\\
    \end{split}
\end{equation}

\subsubsection{Double Deep Q-learning}

For the setting of Deep Q-learning, the network parameter $\mathbf{\theta}$ is optimized by minimizing the loss function $L(\mathbf{\theta})$, which is defined as the difference between the predicted action-value $Q$ and the target action-value $Y^Q$. $\theta$ can be updated with a learning rate $\alpha$, as shown in Equation \ref{equ_dqnL}.
\begin{equation}
    \label{equ_dqnL}
    \begin{split}
        Y_t^Q  & = R_{t+1} + \gamma \max_a Q(S_{t+1}, a | \mathbf{\theta_t})\\
        L(\mathbf{\theta}) & = \left(Y_t^Q -  Q(S_t, A_t | \mathbf{\theta}_t ) \right)^2\\
        \mathbf{\theta}_{t+1} & = \mathbf{\theta}_{t} + \alpha \frac{\partial L(\mathbf{\theta})}{\partial \mathbf{\theta}}\\
    \end{split}
\end{equation}

For the Double Deep Q-learning setting, the target action-value $Y^Q$ is revised according to another target Q-network $Q^{'}$ with parameter $\mathbf{\theta}'$:
\begin{equation}
    \label{equ_ddqn}
    \begin{split}
        Y_t^Q  & = R_{t+1} + \gamma Q(S_{t+1}, \arg\max_a Q(S_{t+1}, a | \mathbf{\theta}_t) | \mathbf{\theta}_t')
    \end{split}
\end{equation}

During the training procedure, technologies such as $\epsilon$-greedy approach \cite{humanlevel_drl} and the prioritized experience replay approach \cite{per} can be applied to improve the training performance.

\subsubsection{Hierarchical Reinforcement Learning}

For the HRL model \cite{hdrl} with sequential sub-goals, a meta controller $Q^1$ generates the sub-goal $g$ for the following steps and a controller $Q^2$outputs the actions based on this sub-goal until the next sub-goal is generated by the meta controller.
\begin{equation}
    \label{equ_hq}
    \begin{split}
        Y_{t}^{Q^{1}}  & = \sum_{t'=t+1}^{t+1+N} R_{t'} + \gamma \max_g Q(S_{t+1+N}, g | \mathbf{\theta_t}^{1})\\
        Y_t^{Q^2}  & = R_{t+1} + \gamma \max_a Q(S_{t+1}, a | \mathbf{\theta_t^2}, g)\\
    \end{split}
\end{equation}


\section{Methodology}

In this section we present our proposed model, which is a hierarchical RL network with an explicit attention model, hybrid reward mechanism and a hierarchical prioritized experience replay training schema. We will refer to this model as Hybrid HRL throughout the paper.

\subsection{Hierarchical RL with Attention}

\begin{figure}[!t]
  \centering
  \includegraphics[width=\columnwidth]{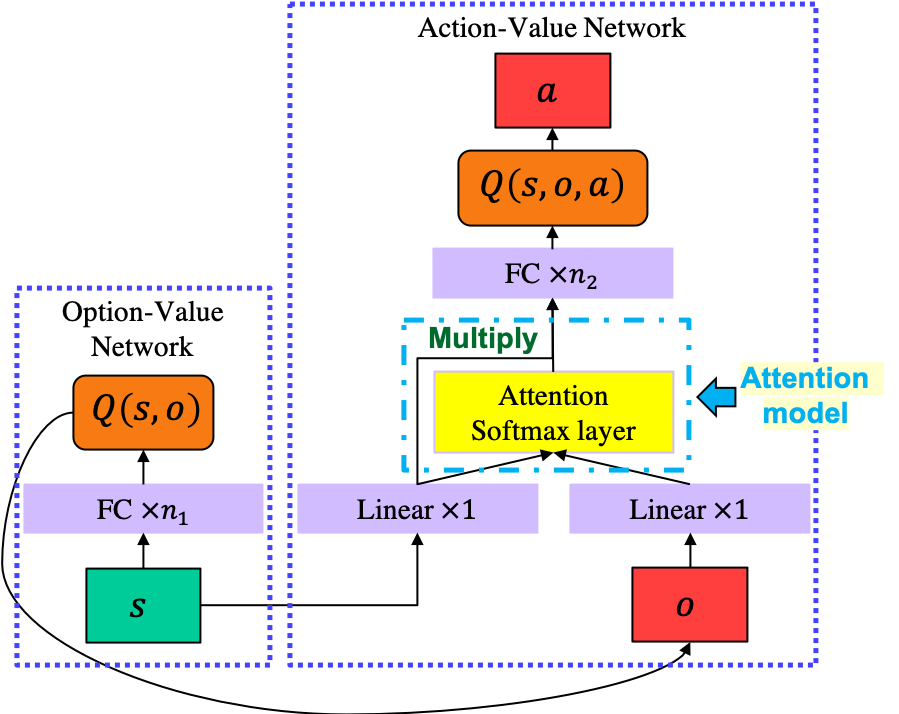}
  \caption{Hierarchical RL Option and Action Q-Network. FC stands for a fully connected layer. Within all the FC layers, \textit{Linear} activation functions are used to generate last layers in both Option-Value and Action-Value networks. For the rest of the layers, \textit{ReLu} activation functions are applied.}
  \label{fig_network}
\end{figure}

Hierarchical structures based on RL can be applied to learn a task with multiple sub-goals. For a hierarchical structure with two levels, an option set $\mathit{O}$ is assigned to the first level, whose object is to select among sub-goals. The weight $\mathbf{\theta}_t^{o} $ is updated according to Equation \ref{equ_qo}.
\begin{equation}
    \label{equ_qo}
    \begin{split}
        O_{t+1}^*  & = \arg\max_o Q^o(S_{t+1}, o | \mathbf{\theta}_t^o)\\
        Y_t^{Q^o}  & = R_{t+1}^o + \gamma Q^o(S_{t+1}, O_{t+1}^* | \mathbf{\theta}_t^{o'})\\
        L(\mathbf{\theta}^o) & = \left(Y_t^{Q^o} -  Q^o(S_t, O_t | \mathbf{\theta}_t^{o} ) \right)^2\\
    \end{split}
\end{equation}

After selecting an option $o$, the corresponding action set $\mathit{A^o}$ represents the action candidates that can be executed on the second level of the hierarchical structure with respect to the selected option $o$. Some previous work proposed the Hierarchical Markov Decision Process (MDP), which shares the state set $\mathit{S}$ among different hierarchical levels during the MDP or designs different states for changing sub-goals and applies initial and terminating condition sets to transfer from one state set to another.

In many situations, the portion of the state set and the amount of abstraction needed to choose actions at different levels of this hierarchy can vary widely. In order to avoid designing a myriad of state representations corresponding to each hierarchy level and sub-goal, we share one state set $\mathit{S}$ for the whole hierarchical structure. Meanwhile, an attention model is applied to define the importance of each state element $I(s, o)$ with respect to each hierarchical level and sub-goal and then use these weights to reconstruct the state $s^I$. The weight $\mathbf{\theta}_t^{a} $ is updated according to Equation \ref{equ_qa}.
\begin{equation}
    \label{equ_qa}
    \begin{split}
        A_{t+1}^*  & = \arg\max_a Q^a(S_{t+1}^I, O_{t+1}^*, a | \mathbf{\theta}_t^a)\\
        Y_t^{Q^a}  & = R_{t+1}^a + \gamma Q^a(S_{t+1}^I, O_{t+1}^*, A_{t+1}^* | \mathbf{\theta}_t^{o'})\\
        L(\mathbf{\theta}^a) & = \left(Y_t^{Q^a} -  Q^a(S_t^I, O_t, A_t | \mathbf{\theta}_t^a ) \right)^2\\
    \end{split}
\end{equation}

When implementing the attention-based HRL, we construct the option network and the action network (Figure \ref{fig_network}), which includes the attention mechanism as a $\mathit{softmax}$ layer in the action-value network $Q^a$.

\subsection{Hybrid Reward Mechanism}

For a sequential sub-goals HRL model \cite{hdrl}, the reward function is designed separately for the sub-goals and main task. The extrinsic meta reward is responsible for the option-level task, and meanwhile the intrinsic reward is responsible for the action-level sub-goals. For HRL with parameterized actions \cite{paramaction}, an integrated reward is designed to evaluate both option-level and action-level together.

\begin{figure}[!t]
  \centering
  \includegraphics[width=0.6\columnwidth]{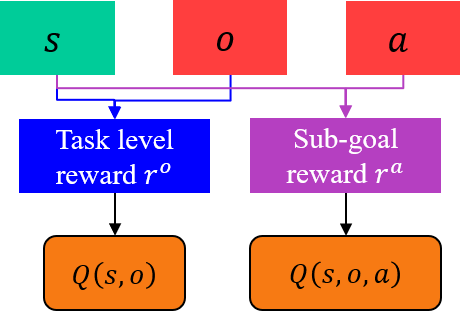}
  \caption{Hybrid Reward Mechanism}
  \label{fig_hybrid_reward}
\end{figure}

In our work, instead of generating one reward function which is applied to evaluate the final outputs coming from both options and actions in one step together, we designed a reward mechanism which can evaluate the goodness of option and action separately during the learning procedure. As a result, a hybrid reward mechanism is introduced so that: 1) the algorithm gets the information of which reward function should be triggered to get rewards or penalties; 2) meanwhile, a positive reward which benefits both option reward and action reward occurs if and only if the whole task and the sub-goals in the hierarchical structure have all been completed. Figure \ref{fig_hybrid_reward} demonstrates the idea for the hybrid reward mechanism.

\subsection{Hierarchical Prioritized Experience Replay}

In \cite{per} the authors propose a framework for more efficiently replaying experience during the training process in DQN so that the stored transitions $\left\{s, a, r, s'\right\}$ with higher TD-error in the previous training iteration result in a higher probability of being selected in the mini-batch for training during the current iteration. However, in the HRL structure, the rewards received from the whole system not only rely on the current level, but also are affected by the interactions among different hierarchical levels.

For the transitions $\left\{s, o, a, r^o, r^a, s'\right\}$ stored during the HRL process, the central observation is that if the output of the option-value network $o$ is chosen wrongly due to high error between predicted option-value $Q^o$ and the targeted option-value $r^o + \gamma Q^o(s', o')$, then the success or failure of the corresponding action-value network is inconsequential to the current transition. As a result, we propose a hierarchical prioritized experience replay (HPER) in which the priorities in the option-level are based on error directly and the priorities in the lower level are based on the difference between errors coming from two levels. Higher priority is assigned to the action-level experience replay if the corresponding option-level has lower priority. According to Equations \ref{equ_qo} and \ref{equ_qa}, the transition priorities for option and action level are given in Equation \ref{equ_pri}.
\begin{equation}
    \label{equ_pri}
    \begin{split}
        p^o & = \left | Y^{Q^o} -  Q^o(S, O | \mathbf{\theta}^{o} ) \right|\\
        p^a & = \left | Y_t^{Q^a} -  Q^a(S_t^I, O_t, A_t | \mathbf{\theta}_t^a ) \right| - p^o
    \end{split}
\end{equation}

Based on the aforementioned approaches, the Hybrid HRL is shown in Algorithm \ref{algo_haddqn}, \ref{algo_hrm} and \ref{algo_hper}.

\begin{algorithm}[!t]\footnotesize
    \caption{Hierarchical RL with Attention State}
    \label{algo_haddqn}
    \begin{algorithmic}[1]
    \Procedure{HRL-AHR()}{}
        \State Initialize option and action network $Q^o$,  $Q^a$ with weights $\theta^{o}$, $\theta^{a}$ and the target option and action network ${Q}^{o'}$, ${Q}^{a'}$ with weights $\theta^{o'}$,  $\theta^{a'}$. 
        \State Construct an empty replay buffer $\mathbf{B}$ with max memory length $l_B$.
        \For {$e \gets 0$ to $E$ training epochs}
            \State Get initial states $s_0$.
            \While {$s$ is not the terminal state}
                \State Select option $O_t = \arg\max_{o} Q^o (S_t, o)$ based on $\epsilon$-greedy. $O_t$ is the selected sub-goal that the lower-level action will execute. 
                \State Apply attention model to state $S_t$ based on the selected option $O_t$: $S_t^I = I(S_t, O_t)$.
                \State Select action $A_t = \arg\max_{a} Q^a (S_t^I, O_t, a) $ based on $\epsilon$-greedy.
                \State Execute $A_t$ in simulation to get $S_{t+1}$.
                \State $R^o_{t+1}, R^a_{t+1} = HybridReward(S_t, O_t, A_t)$.
                \State Store transition $T$ into $\mathbf{B}$: $T = \left\{S_t, O_t, A_t, R^o_{t+1}, R^a_{t+1}, S_{t+1} \right\}$.
            \EndWhile
            \State Train the buffer $ReplayBuffer(e)$.
            \If {$e$ mod $n == 0$}
                \State Test without action exploration with the weights from training results for $n$ epochs and save the average rewards.
            \EndIf
        \EndFor
    \EndProcedure
    \end{algorithmic}
\end{algorithm}

\begin{algorithm}[!t]\footnotesize
    \caption{Hybrid Reward Mechanism}
    \label{algo_hrm}
    \begin{algorithmic}[1]
    \Procedure{HybridReward()}{}
        \State Penalize $R_t^o$ and $R_t^a$ for regular step penalties (e.x.: time penalty).
        \For {$\delta$ in sub-goals candidates}
            \If {$\delta$ fails}
                \If {option $o_t == \delta$}
                    \State Penalize option reward $R_t^o$
                \Else
                    \State Penalize action reward $R_t^a$
                \EndIf
            \EndIf
        \EndFor
        \If {task success (all $\delta$ success)}
            \State Reward both $R_t^o$ and $R_t^a$.
        \EndIf
    \EndProcedure
    \end{algorithmic}
\end{algorithm}

\begin{algorithm}[!t]\footnotesize
    \caption{Hierarchical Prioritized Experience Replay}
    \label{algo_hper}
    \begin{algorithmic}[1]
    \Procedure{ReplayBuffer($e$)}{}
        \State mini-batch size $k$, training size $N$, exponents $\alpha$ and $\beta$.
        \State Sample $k$ transitions for option and action mini-batch:\[MB^g \sim P^g = \frac{{p^{g}}^\alpha}{\sum_0^{l_B} {p_i^{g}}^\alpha}, \quad g \in \left\{ o, a\right\}\]
        \State Compute importance-sampling weights: \[w^g = \frac{\left[N \cdot P^g\right]^{-\beta}}{\max_i w_i^g}, \quad g \in \left\{ o, a\right\}\]
        \State Update transition priorities: \[p^o = \left | Y_t^{Q^o} -  Q^o(S_t, O_t | \mathbf{\theta}_t^{o} ) \right|\]\[p^a  = \left | Y_t^{Q^a} -  Q^a(S_t^I, O_t, A_t | \mathbf{\theta}_t^a ) \right| - p^o\]
        \State Adjust the transition priorities to be greater than 0: $p^a = p^a - min(p^a)$.
        \State Perform gradient descent to update $\theta^{g}_t =  \theta^{g}_t + \alpha \frac{\partial L(\mathbf{\theta^g})}{\partial \mathbf{\theta^g}}$ according to sample weights $w^g$, $g \in \left\{ o, a\right\}$.
        \State Update target networks weights $\theta^{g'} = \theta^{g}$, $g \in \left\{ o, a\right\}$.
    \EndProcedure
    \end{algorithmic}
\end{algorithm}

\section{Experiment}

In this section, we apply the proposed algorithm to the behavior planning of a self-driving car and make comparisons with competing methods.

\begin{table*}[!t]
\caption{Results comparisons among different behavior policies}
\label{table_result}
\begin{center}
\begin{tabular}{|c||c c|c|c c |c c c c|}
\hline
& \multicolumn{2}{c|}{Rewards} & \multirow{2}{*}{Step} & \multicolumn{2}{c |}{Step Penalty} & \multicolumn{4}{ c |}{ Performance Rate}\\
\cline{2-3}\cline{5-10}
& Option Reward $r^o$ &  Action Reward  $r^a$  & & Unsmoothness & Unsafe & Collision & Not Stop & Timeout & Success\\
\hline
Rule 1 & -36.82 & -9.11 & 112 & 0.38 & 8.05 & 18\% & 82\% & 0\% & 0\%\\
Rule 2 & -28.69 & 0.33 & 53 & 0.32 & 6.41 & 89\% & 0\% & 0\% & 11\% \\
Rule 3 & 26.42 & 13.62 & 128 & 0.54 & 13.39 & 31\% & 0\% & 0\% & 69\%\\
Rule 4 &  40.02 & 17.20 & 149 & 0.58 & 16.50 & 14\% & 0\% & 0\% & 86\%\\
Hybrid HRL &  43.52 & 28.87 & 178 & 5.32 & 1.23 & 0\% & 7\%  & 0\% & 93\% \\
\hline
\end{tabular}
\end{center}
\end{table*}

\subsection{Scenario}

\begin{figure}[!t]
  \centering
  \includegraphics[width=\columnwidth]{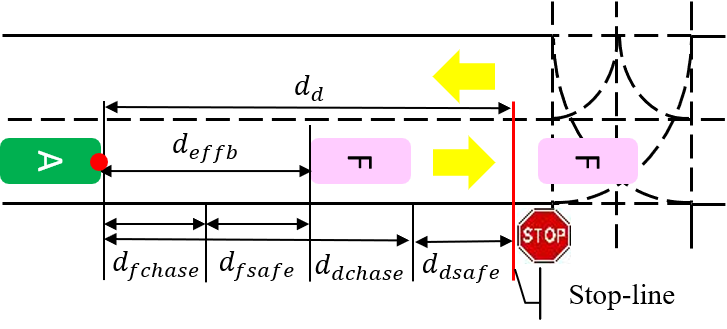}
  \caption{Autonomous vehicle (green box with $A$) approaching stop-sign intersection}
  \label{fig_scenario}
\end{figure}

We tested our algorithm in MSC's VIRES VTD, which is a complete simulation tool-chain for driving applications \cite{vtd}. We designed a task in which an autonomous vehicle (green box with $A$) intends to stop at the stop-line behind a random number of front vehicles (pink boxed with $F$) which have random initial positions and behavior profiles (see Figure \ref{fig_scenario}). The two sub-goals in this scenario are designed as \textit{STOP AT STOP-LINE} (\textit{SSL}) and \textit{FOLLOW FRONT VEHICLE} (\textit{FFV}).

\subsection{Transitions}

\subsubsection{State}

The state which is used to formulate the hierarchical deep reinforcement learning includes the information of the ego car, which is useful for both sub-goals, and the related information that is needed for each sub-goal.
\begin{equation}
    \label{equ_state}
    s = \left[v_e, a_e, j_e, d_{f}, v_f, a_f, d_{fc}, \frac{d_{fc}}{d_{fs}}, d_d, d_{dc},  \frac{d_{dc}}{d_{ds}} \right]
\end{equation}

Equation \ref{equ_state} describes our state space where $v_e$, $a_e$ and $j_e$ are respectively the velocity, acceleration and jerk of the ego car, while $d_f$ and $d_d$ denote the distance from the ego car to the nearest front vehicle and the stop-line, respectively. A safety distance parameter is introduced as a nominal distance behind the target object which can improve safety due to different sub-goals.

\begin{equation}
    \label{equ_dis}
    \begin{split}
        & d_{fs} = \max \left(\frac{v_e^2 - v_f^2}{2 a_{max}}, d_{0}\right), \quad d_{fc}  = d_f - d_{fs} \\
        & d_{ds} = \frac{v_e^2}{2 a_{max}}, \quad d_{dc}  = d_d - d_{ds}
    \end{split}
\end{equation}

Here $a_{max}$ and $d_0$ denote the ego car's maximum deceleration and minimum allowable distance to the front vehicle, respectively, and $d_{fc}$ and $d_{dc}$ are the distances that can be chased by the ego car (distances to the front vehicle minus safety distance of the target). The initial positions of front vehicles and ego car are randomly selected.

\subsubsection{Option and Action}

The option network in the scenario outputs the selected sub-goal: \textit{SSL} or \textit{FFV}. Then, according to the option result, the action network generates the throttle or brake choices.

\subsubsection{Reward Functions}

Assume that for one step, the selected option is denoted as $o$, $o \in \left\{d, f\right\}$. The reward function is given by:

For each step:
\begin{itemize}
    \item Time penalty: $- \sigma_1$.
    \item Unsmoothness penalty if jerk is too large: $-\mathbb{I}_{j_e > 1.} \sigma_2$.
    \item Unsafe penalty: $ - \mathbb{I}_{d_{dc} < 0} \exp(-\frac{d_{dc}}{d_{ds}}) - \mathbb{I}_{d_{fc} < 0} \exp(-\frac{d_{fc}}{d_{fs}})$.
\end{itemize}

 For the termination conditions:
 \begin{itemize}
    \item Collision penalty: $- \mathbb{I}_{d_f = 0.} \sigma_3 $.
    \item Not stop at stop-line penalty: $-\mathbb{I}_{d_d = 0.} v_e^2 $.
    \item Timeout: $- \mathbb{I}_{\text{timeout}}  d_d^2 $.
    \item Success reward: $\mathbb{I}_{d_d = 0., v_e = 0}\sigma_4$
\end{itemize}
where $\sigma_k$ are constants. $\mathbb{I}_c$ are indicator functions. $\mathbb{I}_c = 1$ if and only if $c$ is satisfied, otherwise $\mathbb{I}_c = 0$.

Assume that for one step, the selected option is denoted as $o$, $o \in \left\{d, f\right\}$ and the unselected option is $o^{-}$, $o^{-} \in \left\{f, d\right\}$:
\begin{equation}
    \begin{split}
        sr & = - \sigma_1 - \mathbb{I}_{\text{timeout}}  d_d^2 + \mathbb{I}_{d_d = 0., v_e = 0}\sigma_4 \\
        r^{option}  & =  sr - \mathbb{I}_{d_{o^{-}c} < 0} \exp(-\frac{d_{o^{-}c}}{d_{o^{-}s}}) - \mathbb{I}_{d_{o^{-}} = 0.} v_e^2 \\
        r^{action}  & = sr - \mathbb{I}_{j_e > 1.} \sigma_2 - \mathbb{I}_{d_{oc} < 0} \exp(-\frac{d_{oc}}{d_{os}}) - \mathbb{I}_{d_o = 0.} \sigma_3\\ 
        r^{task} & = sr  - \mathbb{I}_{d_{dc} < 0} \exp(-\frac{d_{dc}}{d_{ds}}) - \mathbb{I}_{d_{fc} < 0} \exp(-\frac{d_{fc}}{d_{fs}}) \\
        & \quad -\mathbb{I}_{j_e > 1.} \sigma_2- \mathbb{I}_{d_f = 0.} \sigma_3 -\mathbb{I}_{d_d = 0.} v_e^2 
    \end{split}
\end{equation}
where $sr$ represents the portion of the reward common to $r^{option}$, $r^{action}$ and $r^{task}$. 

For comparison, we also formulate the problem without considering a hierarchical model via Double DQN. Then $r^{task}$ denotes the reward for achieving the task in this flattened action space.

\subsection{Results}

We compare the proposed algorithm with four rule-based algorithms and some traditional RL algorithms mentioned before. Table \ref{table_result} shows the quantitative results for testing the average performance of each algorithm over 100 cases.

\begin{figure}[!t]
  \centering
  \includegraphics[width=\columnwidth]{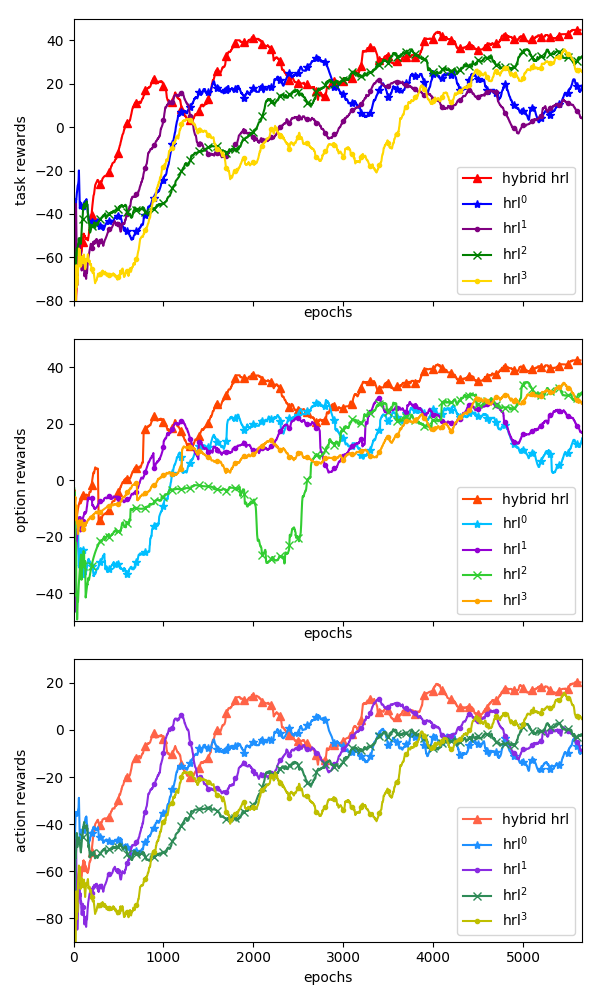}
  \caption{Training results}
  \label{fig_comparison}
\end{figure}

\begin{table}[!t]
\caption{Different HRL-based policies}
\label{table_hrl_candidate}
\begin{center}
\begin{tabular}{|c||c c c|}
\hline
& Hybrid Reward & Hierarchical PER & Attention Model\\
\hline
HRL$^0$ & $\times$ & $\times$ & $\times$ \\
HRL$^1$ & $\surd$ & $\times$ & $\times$  \\
HRL$^2$ & $\surd$ & $\surd$ & $\times$ \\
HRL$^3$ & $\surd$ & $\times$ & $\surd$  \\
Hybrid HRL & $\surd$ & $\surd$ & $\surd$ \\
\hline
\end{tabular}
\end{center}
\end{table}

The competing methods include:
\begin{itemize}
    \item Rule 1: stick to the option \textit{Follow Front Vehicle} (\textit{FFV}).
    \item Rule 2: stick to the option \textit{Stop at Stop-line} (\textit{SSL}).
    \item Rule 3: if $d_d > (d_f + car\_length)$, select \textit{FFV}, w/o \textit{SSL}.
    \item Rule 4: if $d_f > d_{fc}$, select \textit{FFV}, w/o \textit{SSL}.
    \item Table \ref{table_hrl_candidate} shows the explanations of different HRL-based algorithms whose results are shown in Figure \ref{fig_comparison}. 
\end{itemize}

Figure \ref{fig_comparison} compares the Hybrid HRL method with different setup of HRL algorithms. The results show that the hybrid reward mechanism can perform better with the help of hierarchical PER approach.

\begin{figure}[!t]
  \centering
  \includegraphics[width=\columnwidth]{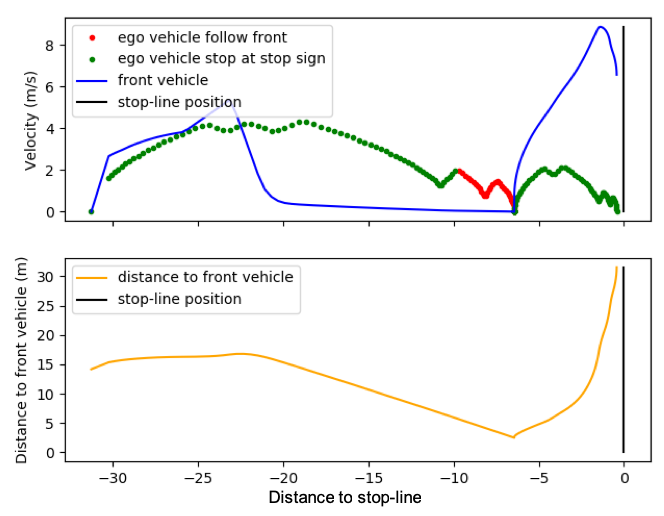}
  \caption{Velocities of ego car and front vehicles}
  \label{fig_v_d1}
\end{figure}

Figure \ref{fig_v_d1} depicts a typical case of the relative speed and position of the ego vehicle with respect to the nearest front vehicle as they both approach the stop-line. In the bottom graph we see the ego vehicle will tend to close the distance to the front vehicle until a certain threshold (about 5 meters) before lowering its speed relative to the front vehicle to allow a certain buffer between them. In the top graph we see that during this time the front vehicle begins to slow rapidly for the stop-line at around 25 meters out before taxing to a stop. Simultaneously, the ego vehicle opts to focus on stopping for the stop-line until it's within a certain threshold of the front vehicle, at which point it will attend to the front vehicle instead. Finally, after a pause the front vehicle accelerates through the stop-line and at this point the ego vehicle immediately begins focusing on the stop sign once again as desired.

\begin{figure}[!t]
  \centering
  \includegraphics[width=\columnwidth]{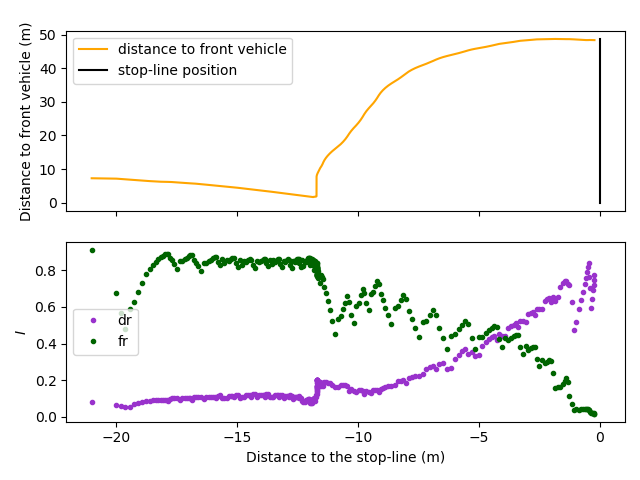}
  \caption{Attention value extracted from the attention layer in the model. $dr$ and $fr$ are $\frac{d_{dc}}{d_{ds}}$ and $\frac{d_{fc}}{d_{fs}}$ in the introduced state, respectively.}
  \label{fig_attention_results}
\end{figure}

\begin{figure}[!t]
  \centering
  \includegraphics[width=\columnwidth]{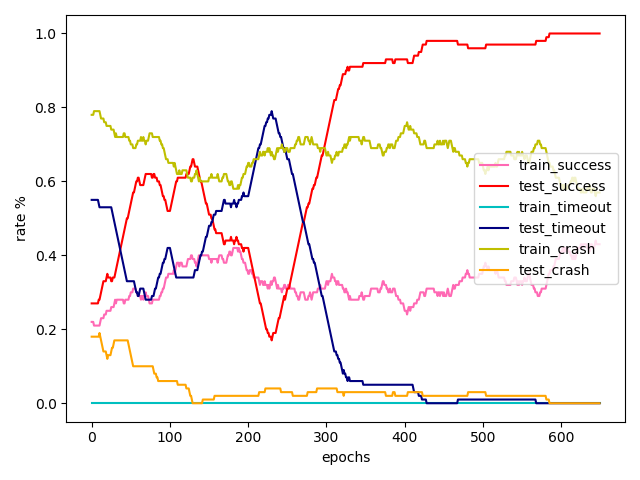}
  \caption{Performance rate of only training to \textit{Follow Front Vehicles} during the training process. Results from training include random actions taken according to explorations. Results from testing show average performance by testing 200 cases based on the trained network after that training epoch.}
  \label{fig_follow_only}
\end{figure}

Figure \ref{fig_attention_results} shows the results extracted from the attention \textit{softmax} layer. Only the two state elements with the highest attentions have been visualized. The upper sub-figure shows the relationship between the distance to the nearest front vehicle (y-axis) and the distance to the stop-line (x-axis). The lower sub-figure is the attention value. When the ego car is approaching the front vehicle, the attention is mainly focused on $\frac{d_{fc}}{d_{fs}}$. When the front vehicle leaves without stopping at the stop-line, the ego car transfers more and more attentions to $\frac{d_{dc}}{d_{ds}}$ during the process of approaching the stop-line.

\begin{figure}[!t]
  \centering
  \includegraphics[width=\columnwidth]{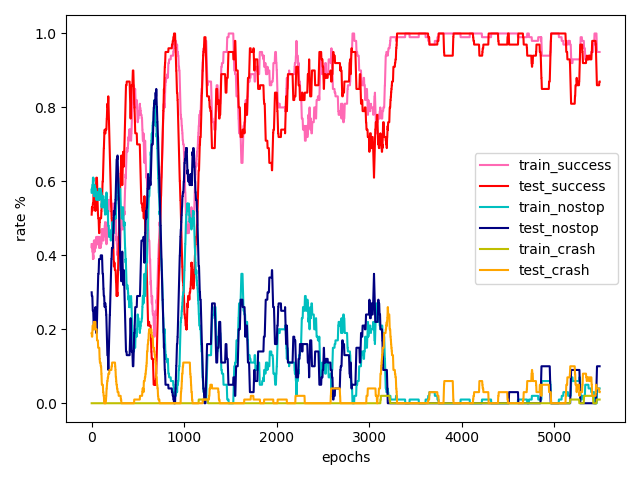}
  \caption{Performance rate of only training to choose the options between \textit{FFV} or \textit{SSL} based on the designed rule-based or trained action-level policies. Results from Test shows average performance by testing 100 cases based on the trained network after that training epoch.}
  \label{fig_option_only}
\end{figure}

\begin{figure}[!t]
  \centering
  \includegraphics[width=\columnwidth]{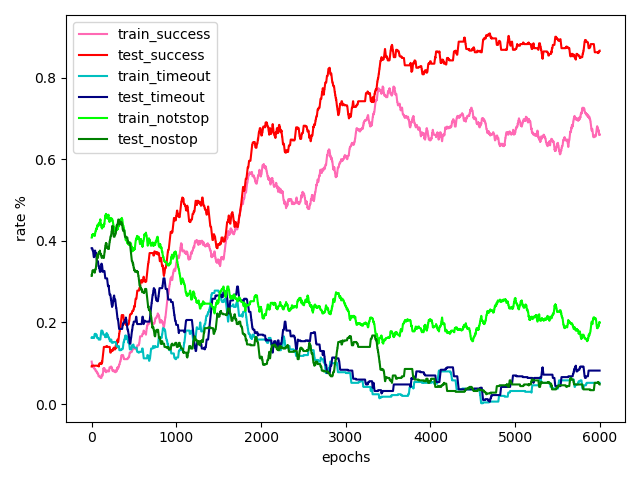}
  \caption{Performance rate of Hybrid HRL training process. Results from testing show average performance by testing 500 cases based on the trained network after that training epoch.}
  \label{fig_hir_success}
\end{figure}

For the scenario of approaching the intersection with front vehicles, one of the methods is to manually design all the rules. Another possibility is to design a rule-based policy of stopping at the stop-line which is relative easy to model. Then we train a DDQN model (see Figure \ref{fig_follow_only} for training process) to be the policy of following front vehicles. Based on these two action-level models, we train another DDQN model (see Figure \ref{fig_option_only} for training process) to be the policy governing which option is needed for approaching the stop-line with front vehicles. During the training process, after every training epoch, the simulation will test 500 epochs without action exploration based on the trained-out network. By applying the proposed hybrid HRL, all the option-level and action-level policies can be trained together (see Figure \ref{fig_hir_success} for training process) and the trained out policy can be separated if the target task only need to achieve one of the sub-goals. For example, the action-value network of \textit{Following Front Vehicle} can be used alone with the corresponding option input to the network. Then, the ego car can follow the front vehicle without stopping at the stop-line.

\section{CONCLUSIONS}

In this paper, we proposed three extensions to hierarchical deep reinforcement learning aimed at improving convergence speed, sample efficiency and scalability over traditional RL approaches. Preliminary results suggest our algorithm is a promising candidate for future research as it is able to outperform a suite of hand-engineered rules on a simulated autonomous driving task in which the agent must pursue multiple sub-goals in order to succeed.

\section*{Acknowledgments}

The authors would like to thank S. Bilal Mehdi of General Motors Research \& Development for his assistance in implementing the VTD simulation environment used in our experiments.

\addtolength{\textheight}{-12cm}   






\bibliographystyle{IEEEtran}
\bibliography{root.bbl}

\end{document}